\documentclass[10pt,twocolumn,letterpaper]{article}

\usepackage{cvpr}
\usepackage{times}
\usepackage{epsfig}
\usepackage{graphicx}
\usepackage{amsmath}
\usepackage{amssymb}
\usepackage{lipsum}
\usepackage{graphicx}
\usepackage[caption=false, font=footnotesize]{subfig}

\newcommand*{\affaddr}[1]{#1} 
\newcommand*{\affmark}[1][*]{\textsuperscript{#1}}
\newcommand*{\email}[1]{\texttt{#1}}

\usepackage[pagebackref=true,breaklinks=true,letterpaper=true,colorlinks,bookmarks=false]{hyperref}

\cvprfinalcopy 

\def\cvprPaperID{3619} 

\ifcvprfinal\fi
\begin{document}

\title{Conditional Adversarial Generative Flow for Controllable Image Synthesis}

\author{%
Rui Liu\affmark[1] \quad Yu Liu\affmark[1] \quad Xinyu Gong\affmark[2] \quad Xiaogang Wang\affmark[1] \quad Hongsheng Li\affmark[1]\\
\affaddr{\affmark[1]CUHK-SenseTime Joint Laboratory, Chinese University of Hong Kong \qquad \affmark[2]Texas A\&M University}\\
\email{ruiliu@cuhk.edu.hk} \quad \email{xy\_gong@tamu.edu}\\
\email{\{yuliu, xgwang, hsli\}@ee.cuhk.edu.hk}\\
}

\maketitle
\thispagestyle{empty}

\begin{abstract}
Flow-based generative models show great potential in image synthesis due to its reversible pipeline and exact log-likelihood target, yet it suffers from weak ability for conditional image synthesis, especially for multi-label or unaware conditions. This is because the potential distribution of image conditions is hard to measure precisely from its latent variable $z$. In this paper, based on modeling a joint probabilistic density of an image and its conditions, we propose a novel flow-based generative model named conditional adversarial generative flow (CAGlow). Instead of disentangling attributes from latent space, we blaze a new trail for learning an encoder to estimate the mapping from condition space to latent space in an adversarial manner. Given a specific condition $c$, CAGlow can encode it to a sampled $z$, and then enable robust conditional image synthesis in complex situations like combining person identity with multiple attributes. The proposed CAGlow can be implemented in both supervised and unsupervised manners, thus can synthesize images with conditional information like categories, attributes, and even some unknown properties. Extensive experiments show that CAGlow ensures the independence of different conditions and outperforms regular Glow to a significant extent. 
\end{abstract}

\section{Introduction}
\label{section1}

Generative adversarial networks (GANs)~\cite{wgan,GAN,lsgan,dcgan} and variational auto-encoders (VAEs)~\cite{Kingma2014} are two types of the most popular generative models due to their solid theoretic foundation and excellent results. Also, the performance of conditional image synthesis by these models improves rapidly with the fast development of deep learning. However, GANs have no explicit encoder to map images into a latent space, which is useful for many downstream tasks while the generated images by VAEs tend to be blurry. These problems remain for conditional versions of these models~\cite{cgan,acgan,CVAE,catgan}. Recently flow-based generative models draw increasing attention due to its natural reversibility of mapping between image space and latent space, exact log-likelihood, and its great potential in image synthesis~\cite{nice,realnvp,FFJORD,glow}. In this work we focus on conditional image synthesis by flow-based generative model.

\begin{figure} 
  \centering
  \subfloat[Inferred\label{tsne1}]{%
        \includegraphics[width=0.499\linewidth]{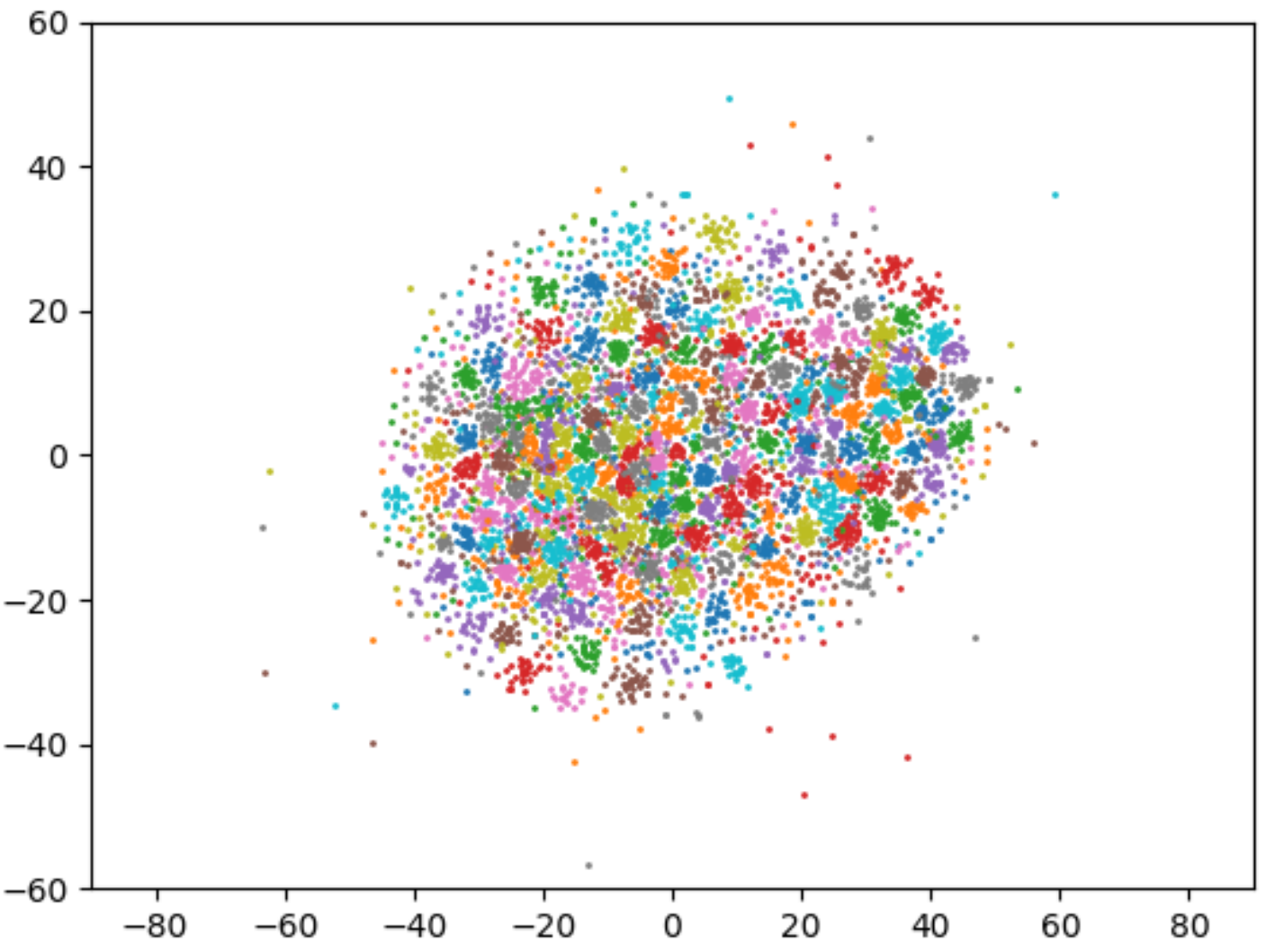}}
    \hfill
  \subfloat[Sampled\label{tsne2}]{%
        \includegraphics[width=0.499\linewidth]{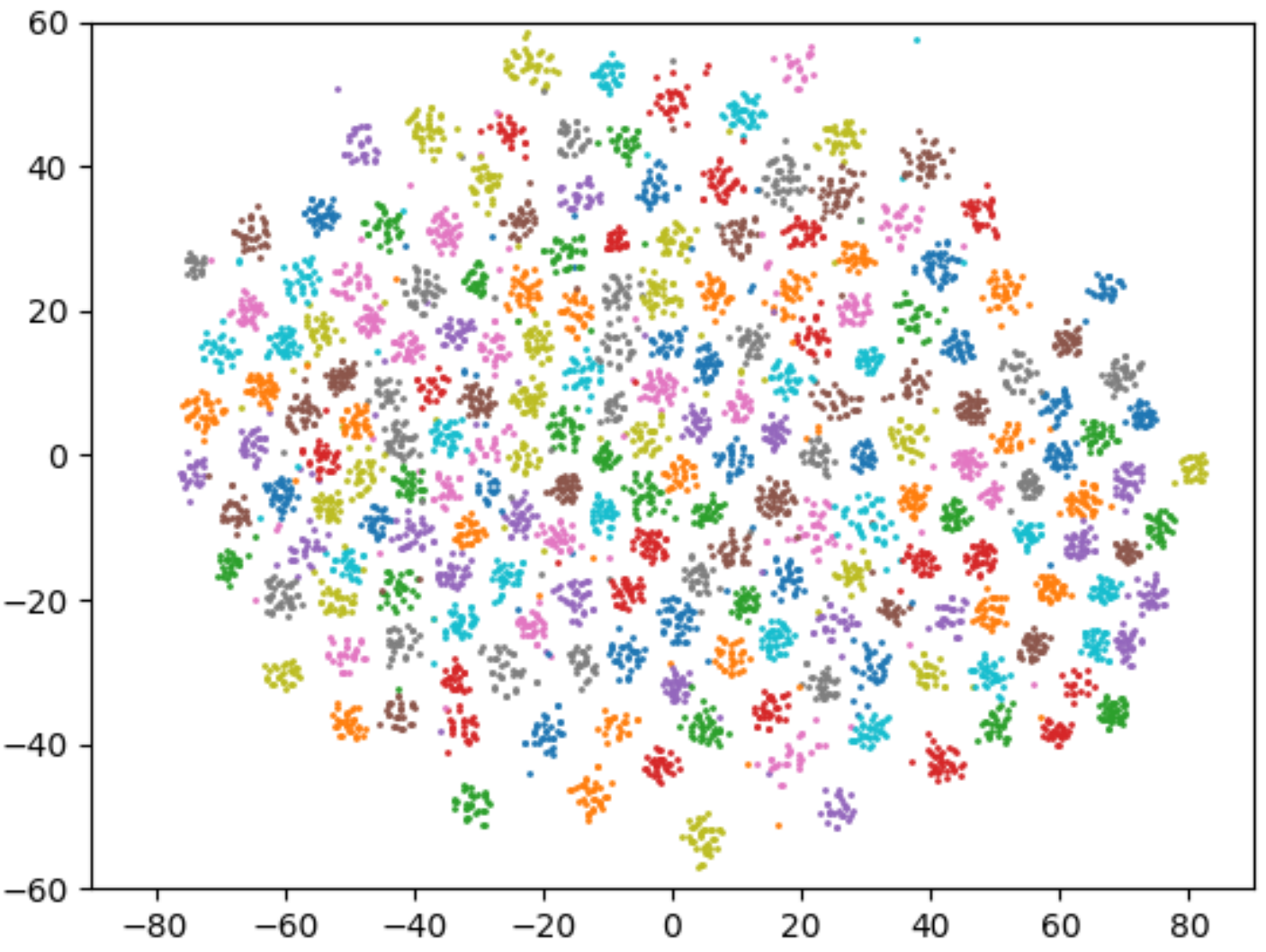}}
  \caption{Barnes-Hut \textit{t}-SNE~\cite{tsne} visualization of $6,000$ latent vectors on $200$ identities of CGlow~\cite{glow}. (a) latent vectors inferred by forward CGlow; (b) randomly sampled latent vectors by inverse CGlow. Best viewed in color.}
  \label{fig:tsne} 
\end{figure}
Unfortunately, conditional image synthesis is a challenging task for flow-based generative models, as these models are forced to have a bijective mapping between the distributions of images and latent vectors according to their definitions~\cite{nice}, which means that their latent dimension must match visible dimension~\cite{goodfellow2016tutorial}. So there is no way to concatenate conditional information with images into the intact model like CGAN~\cite{cgan}, CVAE~\cite{CVAE} and CVAE-GAN~\cite{cvaegan}. Another straight-forward idea is to add a discriminative regularization to the optimization objective like~\cite{DiscoveringHiddenFactors, DisentanglingFactors} with a class dependent prior, as mentioned in the work of original Glow~\cite{glow}. We name this incremental variant of Glow as \textit{CGlow} in this paper. But it tends to fail when meeting complicated conditions, for example, a face dataset with $200$ identities. As shown in Figure~\ref{fig:tsne}, the distribution of real latent vectors inferred by forward CGlow has very close clusters, but the clusters of sampled latent vectors keep far apart and have a large divergence from the real distribution, which leads to artifacts in its generated images, as shown in Figure~\ref{fig:id_attr}. This phenomenon results from that the underlying distribution of image conditions is difficult to measure precisely on the latent space, not to mention some multi-target tasks such as Pose-Invariant Face Recognition~\cite{PosinvSurvey,DRGAN} and Identity-Attribute Disentanglement~\cite{FDGAN, D2AE}. This methods has no way to explore some unknown properties hidden on the latent space either~\cite{infogan}.

To tackle the problems of flow-based generative models mentioned above, we propose a novel conditional flow-based generative model, named as conditional adversarial generative flow (CAGlow). Instead of disentangling representation on the latent space directly, which is a difficult task for flow-based models, this approach learns an effective encoder to map the distribution of conditions into a latent space and builds a tight connection between the real and generated distributions in an adversarial manner. The main contributions of this work are summarized as follows:

\begin{itemize}
  \item We are the first to learn a mapping from conditions to images by using an irreversible encoder to map conditions into the latent space of reversible flow-based models, which can make use of its reversibility to perform controllable image synthesis. 
  \item We also incorporate adversarial networks into the proposed CAGlow, which helps the encoder learn a continuous mapping between condition space and latent space by adversarial training. 
  \item By performing extensive experiments, we testify that CAGlow outperforms the state-of-the-art flow-based model Glow on complex conditions, and this approach can perform image synthesis conditioned on some unknown but interpretable representations learned in an unsupervised fashion.
\end{itemize}

\section{A Review of Flow-based Generative Model and Conditional Image Synthesis}
\label{section2}
Before going deep into the proposed conditional adversarial generative flow, we take a short review of some state-of-the-art generative models and conditional image synthesis models from a probabilistic viewpoint, which acts as a basic theory of our work.

\subsection{Three Basic Generative Models}

There are three basic types of commonly used generative models: generative adversarial networks (GANs)~\cite{GAN}, variational auto-encoders (VAEs)~\cite{Kingma2014} and flow-based generative models (FGMs)~\cite{nice,realnvp,glow}. GAN contains a discriminator and a generator model playing a minimax game. Such a two-player game is actually optimized when they reach the Nash-Equilibrium point, that is, the discriminator can not tell whether an image is real or not. Many following works improve generative adversarial networks by better loss, training skills and evaluating metrics~\cite{wgan,NIPS2017_7240,GANlandscape,itp,ProgressGrowGAN,FID,lsgan,NIPS2016_6125}. The objective of VAEs is to maximize the variational lower bound of log-likelihood of target data points. This lower bound is composed of a KL divergence, between the distribution modeled by the encoder and a prior distribution of latent vectors, and the reconstruction loss between the output and input data. Since there are different strengths and weaknesses in these two types of models, many works are proposed to take full advantages of both of them for promoting image synthesis~\cite{vaegan,aae,AVB}.

Unlike the former two models, flow-based generative models build up a series of invertible transformations and directly optimize the negative log-likelihood of data distribution.

\subsection{Flow-based Generative Models}

As mentioned above, flow-based generative models~\cite{nice,realnvp,glow} aim to map the distribution of natural image $p^*(x)$ into a latent prior distribution $p^*(z)$ using a bijective function $F$, that is $z=F(x)$. Because the function $F$ is bijective, $x=F^{-1}(z)$ is valid as well and could be used to generate images. So its objective for maximizing log-likelihood can be formulated by change of variables: 
\begin{equation}
\label{eq:flow}
\begin{aligned}
  \log p^*(x) & = \log p^*(z) + \log \bigg| \det \frac{dF}{dx}\bigg| \\
  & = \log p^*(z) + \sum_{i=1}^{K}{ \log \bigg| \det \frac{dh_{i}}{dh_{i-1}} \bigg| },
\end{aligned}
\end{equation}
where we define $dh_{0}=x$ and $dh_{K}=z$ for conciseness and the scalar $\log | \det dh_{i}/dh_{i-1}|$ is the absolute value of the log-determinant of the Jacobian matrix $ dh_{i}/dh_{i-1} $. Such Jaconbian matrices lie on the design of bijective functions such as affine coupling layers and invertible $1\times1$ convolutions. Please refer to \cite{realnvp} and \cite{glow} for more details.

\subsection{Conditional Image Synthesis}
Mainstream conditional generative models consist of VAEs and GANs. Along the line of VAEs, CVAE~\cite{CVAE} was proposed to extend the traditional VAE to a conditional generative model, which models a conditional distribution and finds the variational lower bound of this distribution following the idea of vanilla VAE. 
Along the other line of GANs, there exist more conditional models with different forms and applications~\cite{biggan,SemanticLayout,pix2pix,text2img,HRsynthesis,stackgan,cyclegan}. To the best of our knowledge, the pioneer work is CGAN~\cite{cgan}, which concatenates noise or images with class labels and then feeds them into the generator for conditional image synthesis. This idea is simple but lacks efficiency when dealing with multi-category classification tasks. Then ACGAN~\cite{acgan} was proposed to tackle such problems by simply presenting an auxiliary classifier for the discriminator. Another amazing work is infoGAN~\cite{infogan}, which learns interpretable and disentangled representation in a totally unsupervised fashion and provides an elegant theory based on maximizing the mutual information between the input latent codes and their observations. 

The Auxiliary Classifier Generative Adversarial Network (ACGAN) is a classical variant of vanilla GAN, whose objective functions are summarized by:
\begin{equation}
\label{eq:acgan}
\begin{aligned}
  \textit{L}_{s} &= \mathbb{E}_{x \sim p^*(x)}[\log D_{\phi}(x)] + \mathbb{E}_{x \sim p_{\theta}(x)}[\log (1-D_{\phi}(x))], \\
  \textit{L}_{c} &= \mathbb{E}_{x\sim p^*(x), c\sim p(c)}[\log p_{\phi}(c|x)] \\
  &  \quad \quad \quad   \quad \quad \quad \quad   + \mathbb{E}_{x \sim p_{\theta}(x), c\sim p(c)} [\log p_{\phi}(c|x)],
\end{aligned}
\end{equation}
where $p^*(x)$ denotes the real distribution of images $x$, $p_{\theta}(x)$ denotes the generated one, and the discriminator $D_{\phi}$ and the classifier $p_{\phi}(c|x) = C_{\phi}(x)$ share the parameters of two-in-one neural networks. The minimax game is optimized by training the generator to maximize $\textit{L}_{c} - \textit{L}_{s}$ and training the discriminator/classifier to maximize $\textit{L}_{c} + \textit{L}_{s}$.

As we know, the objective of GAN is actually to minimize the Jensen-Shannon Divergence between the real and fake distribution~\cite{wgan}. So the objective above can also be described as to maximize:
\begin{equation}
\label{eq:acgan_js}
\begin{aligned}
  -JS(p^*(x)||p_{\theta}(x)) &+  \mathbb{E}_{x\sim p^*(x), c\sim p(c)}[\log p_{\phi}(c|x)] \\
  &   + \mathbb{E}_{x \sim p_{\theta}(x), c\sim p(c)} [\log p_{\phi}(c|x)].
\end{aligned}
\end{equation}

Furthermore, there are some models combining GANs with VAEs to boost the generation performance like~\cite{aae,AVB}. CVAE-GAN~\cite{cvaegan} is a conditional generative model unifying VAEs with GANs. It first encodes images with labels into latent vectors and then exploits the encoded vectors and same labels to generate images conditionally with the help of a real or fake discriminator and an auxiliary classifier. This model shows great potential in dealing with fine-grained classification problems.

Many empirical studies show the generated images from GANs are sharper than those of VAEs, for both unconditional and conditional models~\cite{biggan, ProgressGrowGAN}. However, unlike variational auto-encoders and flow-based generative models, classic GANs have no encoder to map natural images into latent space, which is useful for downstream tasks such as image editing, inpainting and attribute morphing. Furthermore, different from VAEs that optimize the lower bound of maximum likelihood and infer the latent variable approximately, the objective of flow-based generative model is to optimize the exact log-likelihood directly and infer the latent variable without sampling. 

Therefore, in this paper we take full advantages of the latent space of flow-based models by building a continuous mapping from condition space to latent space and capture the targeted distribution precisely by adversarial networks.

\section{Conditional Adversarial Generative Flow}
\label{section3}
\begin{figure*}
    \centering
    \includegraphics[width=0.82\linewidth]{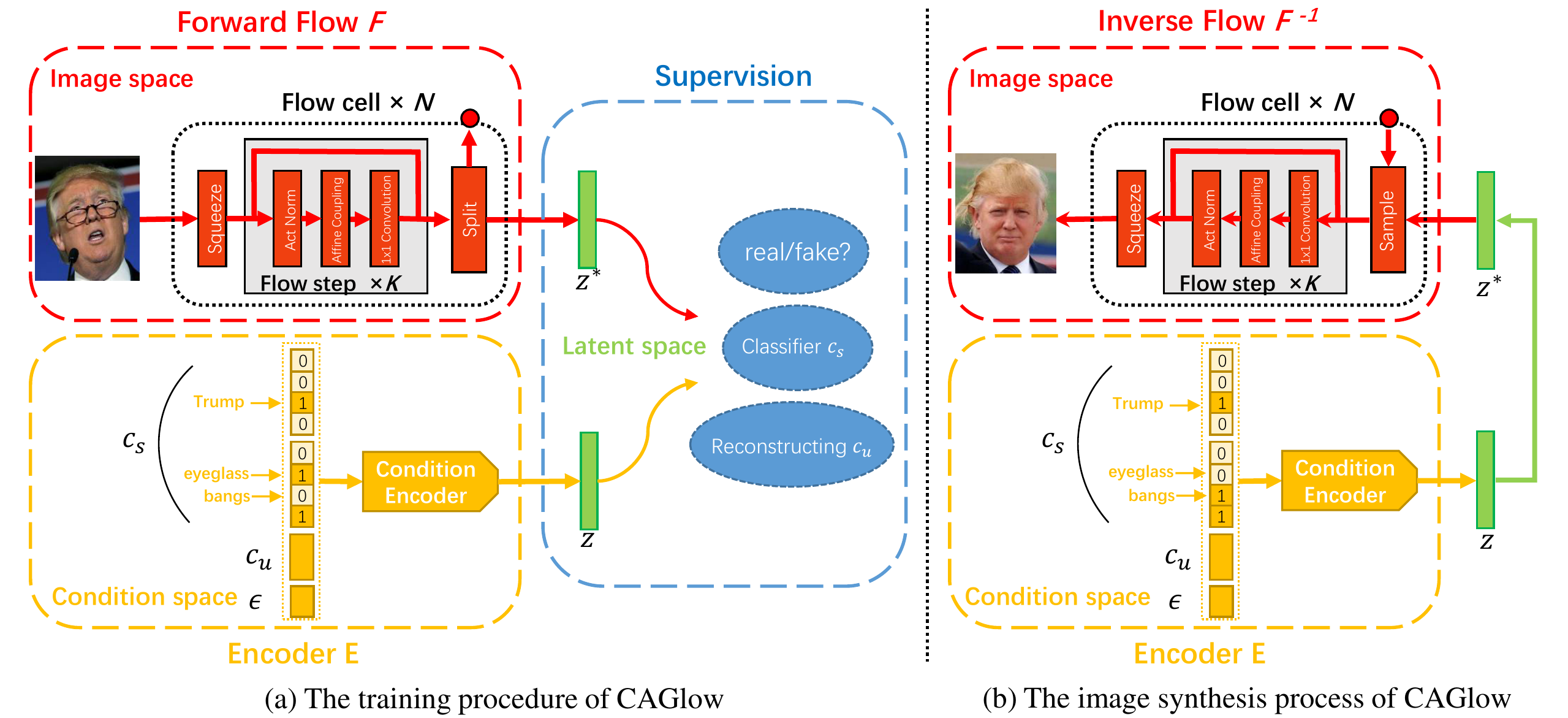}
    \caption{Illustration of the network architecture of the proposed conditional adversarial generative flow. It contains a reversible flow $F$, an encoder $E$, and a supervision block including a discriminator $D_{i}$ distinguishing real vectors from fake ones, a classifier $C$ classifying supervised conditions correctly and a decoder $D_{e}$ reconstructing unsupervised conditions.}
    \label{fig:net}
\end{figure*}
In this section, we introduce the formulation and detailed architecture of our proposed model CAGlow, as shown in Figure~\ref{fig:net}. This model contains a reversible flow, an encoder and a supervision block in general.

\subsection{Formulation}
First, inspired by Eq.(\ref{eq:flow}), we model an image with its conditions as a joint probabilistic distribution and go one step further to obtain the distribution of latent vectors with conditions by a bijective mapping $z=F(x)$:
\begin{equation}
\label{eq:conditional_distribution}
\begin{aligned}
 \log p(x,c_{s}) = \log p(z,c_{s}) + \log \bigg| \det \frac{dF}{dx} \bigg|,
\end{aligned}
\end{equation}
where we let $c_{s}$ denote the conditions under supervision. 

Using Bayesian formula, maximizing equation~\ref{eq:conditional_distribution} is equal to: 
\begin{equation}
\label{eq:bayesian}
\begin{aligned}
  \max \quad & \mathbb{E}_{z\sim p^*(z), c_{s}\sim p(c_{s})}[\log p(c_{s}|z)] \\
  + & \mathbb{E}_{z\sim p^*(z)}[\log p^*(z) + \log \bigg|\det \frac{dF}{dx}\bigg|],
\end{aligned}
\end{equation}
where the prior $p^*(z)$ is modeled by a standard Gaussian distribution. 

We assume there is an unknown other distribution $p(z)$ for latent vectors. According to Gibb's inequality \cite{infotheory}, we find a lower bound for $p^*(z)$:
\begin{equation}
\label{eq:gibbs}
\begin{aligned}
  &\mathbb{E}_{z\sim p^*(z)}[\log p^*(z)] \geq \mathbb{E}_{z\sim p^*(z)}[\log p(z)] \\
  = & \mathbb{E}_{z\sim p^*(z)}[\log p^*(z)] - KL(p^*(z)||p(z)).
\end{aligned}
\end{equation}

Second, we model all the conditions as $ p(\tilde{c}) = p(c_{s}, c_{u})$ where $c_{s}$ denotes the supervised conditions and $c_{u}$ denotes the unsupervised ones. Thus we could use an encoder $E$ to map the conditional information with random noises into a latent distribution $p_{\theta}(z)=E_{\theta}(\tilde{c}, \epsilon)$ where $\epsilon$ denotes random noise.

Using the variational lower bound methods from VAEs~\cite{Kingma2014}, we can find a lower bound for $p(\tilde{c})$ by
\begin{equation}
\label{eq:c_lowerbound}
\begin{aligned}
  &\log p(\tilde{c}) \geq -KL(p_{\theta}(z)||p(z))+\mathbb{E}_{z\sim p_{\theta}(z), \tilde{c}\sim p(\tilde{c})}[\log p(\tilde{c}|z)].
\end{aligned}
\end{equation}

Here we define $p(z)=(p_{\theta}(z)+p^*(z))/2$, so we have $KL(p_{\theta}(z)||p(z))+KL(p^*(z)||p(z))=JS(p_{\theta}(z)||p^*(z))$. Also, we propose a classifier $C$ to classify $z$ from both real and fake distributions. At last, by bringing together all the Eq.(\ref{eq:conditional_distribution}-\ref{eq:c_lowerbound}), we obtain our final objective to maximize:
\begin{equation}
\label{eq:obj}
\begin{aligned}
  & \mathbb{E}_{z\sim p^*(z)}[\log p^*(z) + \log \bigg|\det \frac{dF}{dx}\bigg|] \\
  - & JS(p_{\theta}(z)||p^*(z)) + \mathbb{E}_{z\sim p^*(z), c_{s}\sim p(c_{s})}[\log p(c_{s}|z)] \\
  + & \mathbb{E}_{z\sim p_{\theta}(z), \tilde{c}\sim p(\tilde{c})}[\log p(\tilde{c}|z)].
\end{aligned}
\end{equation}

This objective function could be decomposed into two parts: the first term is the same as the objective of the reversible flow Eq.(\ref{eq:flow}), and the last three terms are very similar to the objective of ACGAN Eq.(\ref{eq:acgan_js}). The difference is that $p(\tilde{c})$ contains both supervised and unsupervised conditional information. Here we assume that they are independent with each other and implement them with a classifier and a decoder, which are illustrated in following part.

\subsection{Network Structure}
Considering our target is to maximize Eq.(\ref{eq:obj}), we would introduce the proposed network structure carefully for achieving this goal. As can be seen in Figure~\ref{fig:net}, the proposed model contains three parts: 1) a reversible multi-scale flow $F$; 2) an encoder $E_{\theta}$; and 3) a supervision block which is a three-in-one neural network including a discriminator $D_{i\phi}$, a classifier $C_{\phi}$ and a decoder $D_{e\phi}$. 

\noindent \textbf{Reversible flow} $F$ builds a bijective mapping between the distributions of natural images and latent vectors using reversible networks formulated as $z=F(x)$ where $z$ has a prior distribution $p^*(z)$. Here we take a standard Gaussian distribution for modeling $z$ and optimize it using maximum likelihood estimation. Specifically, we take the structure of Glow $N\times K$ as our baseline as shown in Figure~\ref{fig:net}. So the loss for reversible flow is 
\begin{equation}
\label{loss:F}
    \mathcal{L}_{F} = - \mathbb{E}_{z\sim p^*(z)}[\log p^*(z) + \log \bigg|\det \frac{dF}{dx} \bigg|].
\end{equation}

Note that samples from $p^*(z)$ are taken as the real data which are fed into the supervision block for further adversarial training, so we take multi-stage training strategy and the first stage is to train a regular Glow model for the following efficient sampling of the latent vectors. An extra advantage of this strategy is that after training, the pretrained Glow model could be used for different tasks by adding different supervised signals on the small supervision blocks, getting rid of the large computation consumption for training many different conditional Glow models on different tasks. 

\noindent \textbf{Encoder} $E_{\theta}$ helps to model the conditional distribution of latent vectors $z$ on conditions $\tilde{c}$. That is,  $p_{\theta}(z)=E_{\theta}(\tilde{c}, \epsilon)$ where $\epsilon$ is from an underlying distribution $p(\epsilon)$ modeled by a standard Gaussian distribution to help $E$ generate diverse samples of latent vectors. $p(\tilde{c})$ is actually modelling the joint distribution for both supervised conditions $c_{s}$ and unsupervised conditions $c_{u}$. Take Figure~\ref{fig:net} as an example, when a face image is fed into the forward flow $F$, its supervised conditions $c_{s}$ containing identity number and attributes like eyeglasses and bangs are fed into the encoder $E$ as one-hot vectors, and simultaneously offer a supervised signal from the top of the classifier $C$. Meanwhile, an unsupervised condition $c_{u}$ and a random noise $\epsilon$ are sampled from their specific distribution and concatenated with the supervised conditions. $c_{u}$ will be decoded from the latent vectors by a decoder $D_{e}$ to enhance its mutual information with $z$. 
According to the objective Eq.(\ref{eq:obj}), We would like to minimize the JS Divergence between this conditional distribution $p_{\theta}(z)$ and the distribution of real latent vectors $p^*(z)$ inferred by the forward flow, with the help of discriminator $D_{i\phi}$. So the loss for the encoder $E_{\theta}$ is:
\begin{equation}
\label{eq:E}
  \mathcal{L}_{E} = - \mathbb{E}_{\epsilon \sim p(\epsilon), \tilde{c} \sim p(\tilde{c})} [\log D_{i\phi}(E_{\theta}(\tilde{c}, \epsilon ))].
\end{equation}

\noindent \textbf{Discriminator} $D_{i\phi}$ aims to distinguish generated latent vectors from real ones inferred by reversible flow correspondingly:
\begin{equation}
\label{eq:D}
  \mathcal{L}_{D_{i}} = -\mathbb{E}_{z \sim p^*(z)} [\log D_{i\phi}(z)] - \mathbb{E}_{z \sim p_{\theta}(z)} [1 - \log D_{i\phi}(z)].
\end{equation}

\noindent \textbf{Classifier} $C_{\phi}$ partly shares the parameters with the discriminator $D_{\phi}$ and outputs different class probabilities by softmax or sigmoid functions. We supervise its training by a cross entropy loss or binary cross entropy loss for different specific tasks. By such a neural network parameterized classifier, we can obtain a class posterior probabilities $q_{\phi}(c_{s}|z)$ of both labeled real vectors and generated ones. The loss could be formulated as:
\begin{equation}
\label{eq:C}
\begin{aligned}
  \mathcal{L}_{C} = &- \mathbb{E}_{z \sim p^*(z), c_{s} \sim p(c_{s})} [\log q_{\phi}(c_{s}|z)] \\
  &- \mathbb{E}_{z \sim p_{\theta}(z),c_{s} \sim p(c_{s})} [\log q_{\phi}(c_{s}|z)].
\end{aligned}
\end{equation}

\noindent \textbf{Decoder} $D_{e\phi}$ partly shares the network parameters with the discriminator and classifier, and it aims to decode the unsupervised conditions from the generated latent vectors for reconstructing them. So the loss for the decoder is:
\begin{equation}
\label{eq:decoder}
  \mathcal{L}_{D_{e}} = - \mathbb{E}_{z \sim p_{\theta}(z), c_{u} \sim p(c_{u})} [\log q_{\phi}(c_{u}|z)],
\end{equation}
where $p(c_{u})$ could be modeled by uniform distribution for continuous codes and binomial distribution for discrete codes. Correspondingly, the loss could be set to mean square error and binary cross entropy loss. 

\subsection{Objective of CAGlow}
We show the designed networks for maximizing equation~\ref{eq:obj}, but in practice, the distribution of real latent vectors and generated ones may not overlap with each other, especially during the early stage of training process, and thus the discriminator can separate them accurately. This phenomenon makes the training process unstable and easy to mode collapse. To overcome this typical but important problem, we propose a pair-wise feature matching regularization strategy, which uses an $L2$ loss between the representation of real and fake data points with same conditions. Let $f(z)$ denote the features of the latent vectors $z$ on the intermediate layer of the network of supervision block, so this pairwise feature matching loss is formulated as: 
\begin{equation}
\label{eq:FM}
  \mathcal{L}_{FM} = \frac{1}{2} ||f(z) - f(z') ||_{2}^{2}.
\end{equation}


So the final goal of our proposed CAGlow is to minimize the loss:
\begin{equation}
\label{eq:loss}
  \mathcal{L} = \sum_{S \in \{F, E, D_{i}, C, D_{e}, FM\}} (\lambda_{S} \mathcal{L}_{S}),
\end{equation}
where the exact loss functions are presented in Eq.(\ref{loss:F}-\ref{eq:FM}). Note that the discriminator $D_{i\phi}$, the classifier $C_{\phi}$ and the decoder $D_{e\phi}$ share most of parameters in the supervision networks except for their output layers. $\mathcal{L}_{D_{i}}$ measures how well the discriminator separates the real and fake vectors and $\mathcal{L}_{C}$ measures how good the classifier at classifying different categories, which can be used directly in downstream tasks like semi-supervised learning. $\mathcal{L}_{D_{e}}$ measures how well the decoder reconstructing the input unsupervised codes, which could be used for unknown properties exploration.

\section{Experiments}
In this section, we would empirically demonstrate the advantage of our proposed approach over some leading baselines to a significant extent. 

\subsection{Implementation Details}
\noindent \textbf{Datasets}. We validate the effectiveness of our proposed model on some publicly accessible datasets. The first dataset is MNIST digits dataset~\cite{mnist} containing $50,000$ training data and $10,000$ test data with classes from number $0$ to $9$. 
The second one is the large-scale face dataset CelebA~\cite{celeba} which contains $202,599$ number of face images with $10,177$ number of identities and $40$ binary attributes annotations per image. For CelebA dataset, we choose a relatively small image size $64$ for evaluation due to the large computation consumption of Glow~\cite{glow}. But the notion is the same for larger image size. 

\noindent \textbf{Networks}. In our experiment, we set the reversible flow network to be a typical setting of Glow $N\times K$. $N$ is the number of cells which contains a \textit{Squeeze} and a \textit{Split} operation for downsampling and dimension reduction. $K$ is the number of steps which contains an affine coupling layer and an invertible $1\times1$ convolution. Please refer to~\cite{realnvp,glow} for details. We set Glow $3\times10$ for MNIST and $3\times32$ for CelebA. In the experiments of MNIST, the encoder and supervision block contain a two fully-connected layers with $64$ hidden neurons. The discriminator, classifier and decoder only share the first layer and output different vectors for calculating their own losses. In the experiments of CelebA, the encoder first embeds identities into a fixed-dimensional latent vector and concatenate it with one-hot vectors of attributes and random noise. Then the vectors pass through one fully-connected layer and three deconvolutional layers with upsampling scale $2$, $2$, $1$ and channnel size $128$, $512$, $48$ respectively. The supervision block contains two stride $2$ convolutional layers with channel size $64$, $128$, followed by four specific fully-connected layers for outputting the probabilities for real or fake, different identities, different attributes and reconstructing unsupervised conditions. 

\noindent \textbf{Baselines}. Since the proposed model is an extension from the state-of-the-art flow-based generative model Glow, we mainly testify the superiority of the proposed model CAGlow to the prior work Glow and its incremental variant CGlow~\cite{glow}.

\begin{figure*}[!h]
  \centering
  \subfloat[CGlow~\cite{glow}\label{id_attr_bad}]{%
      \includegraphics[width=0.4925\linewidth]{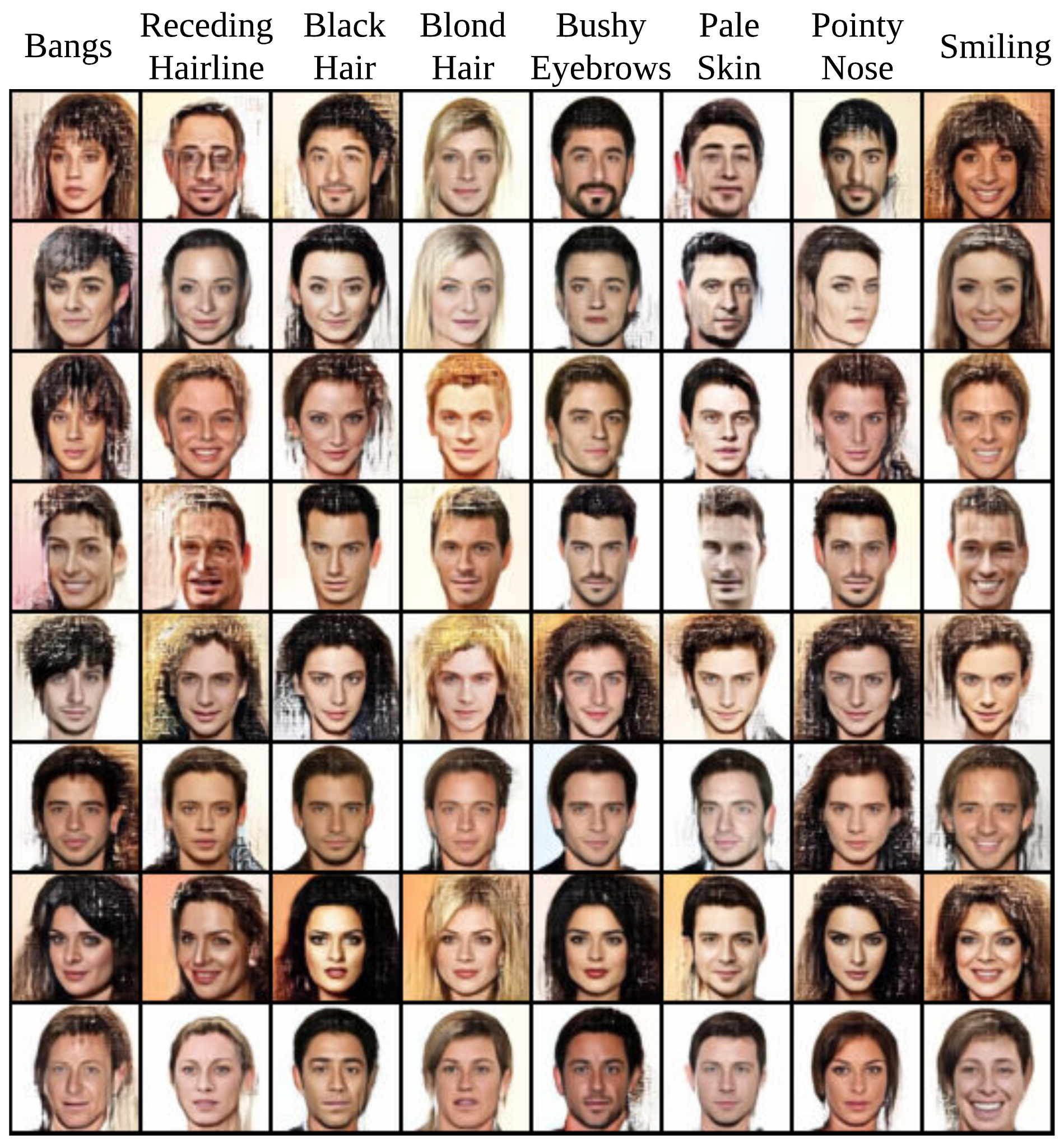}}
    \hfill
  \subfloat[CAGlow\label{id_attr_good}]{%
        \includegraphics[width=0.499\linewidth]{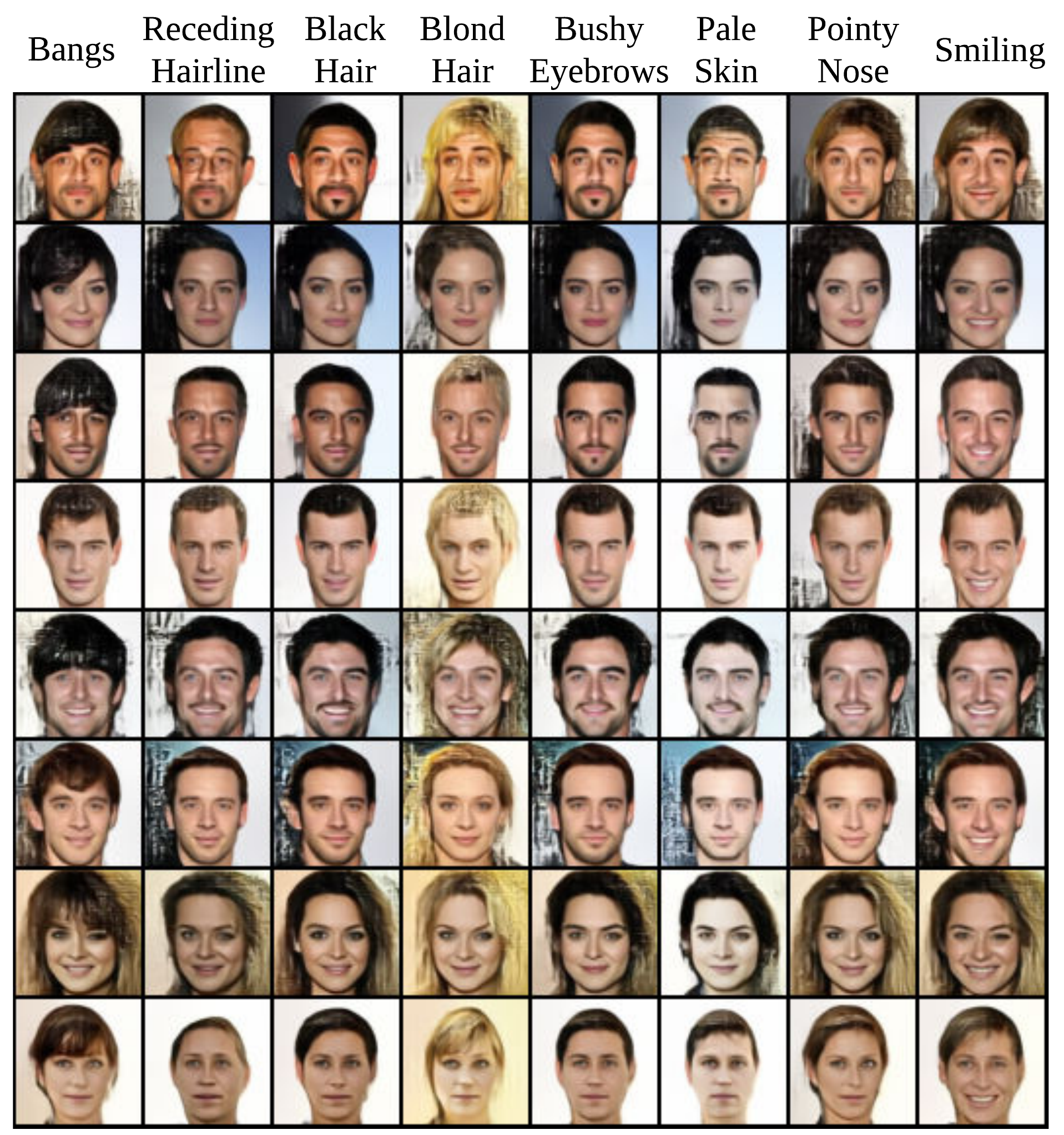}}
  \caption{Conditional image synthesis demonstration. From top to bottom: different people. From left to right: different attributes (specific attribute is annotated above the first row). (a) Images generated by CGlow. Identities and attributes interfere with each other heavily; (b) Images generated by CAGlow with better controllability.}
  \label{fig:id_attr}
\end{figure*}

\subsection{Controllable Image Synthesis}
\label{section42}
\noindent \textbf{Conditional images synthesis} results on different identities and attributes by different approaches are demonstrated in Figure~\ref{fig:id_attr}. We set same identity for each row and same attribute for each column. From Figure~\ref{id_attr_bad}, we could see that the generated images by CGlow are disturbed severely by different identities and attributes. The change of attributes has an adverse impact on the identities and vice versa. In addition, the change of attributes also influences the appearance or disappearance of other attributes in CGlow. Besides, we could see some artifacts in the images generated by CGlow in that the sampling distribution deviates from the real one, as mentioned in Section~\ref{section1}. While the images synthesized by CAGlow avoid such negative effects and show excellent performance under this setting, as shown in Figure~\ref{id_attr_good}. 

\begin{figure*}[!ht]
  \centering
  \subfloat[Glow~\cite{glow}\label{plus1}]{%
      \includegraphics[width=0.49\linewidth]{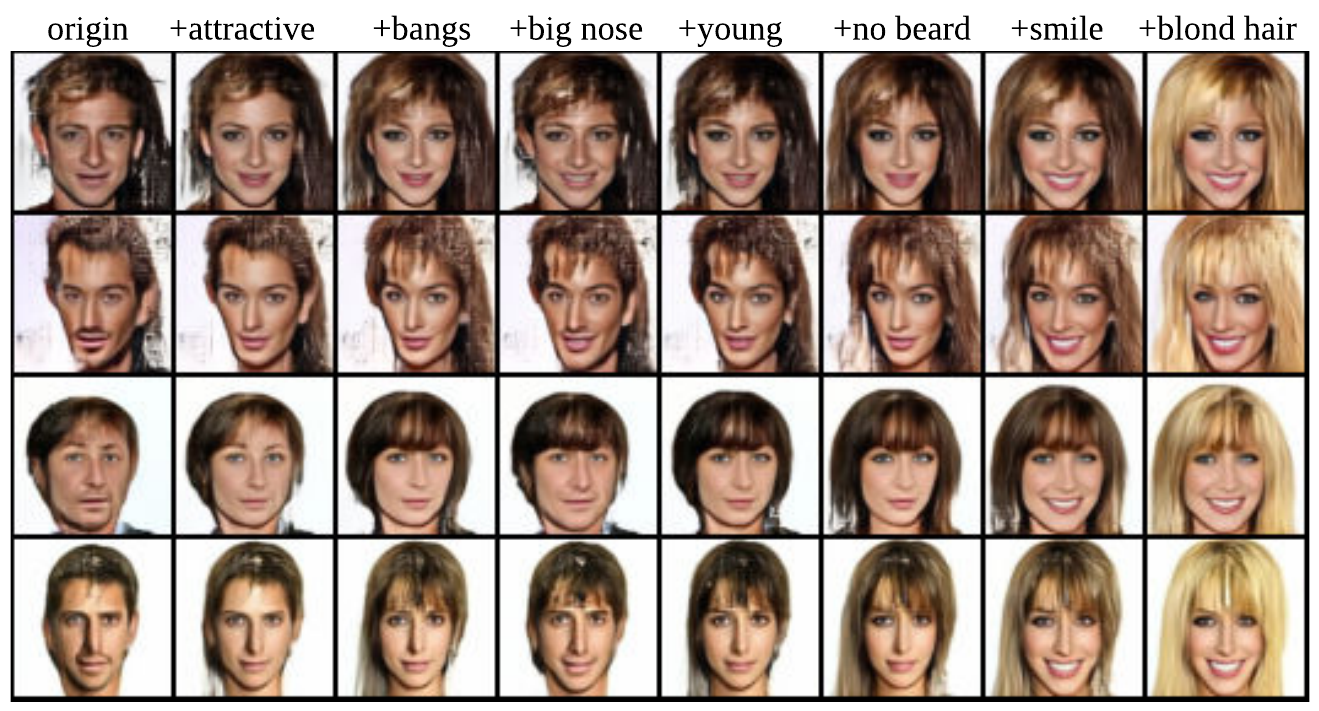}}
    \hfill
  \subfloat[CAGlow\label{plus2}]{%
        \includegraphics[width=0.49\linewidth]{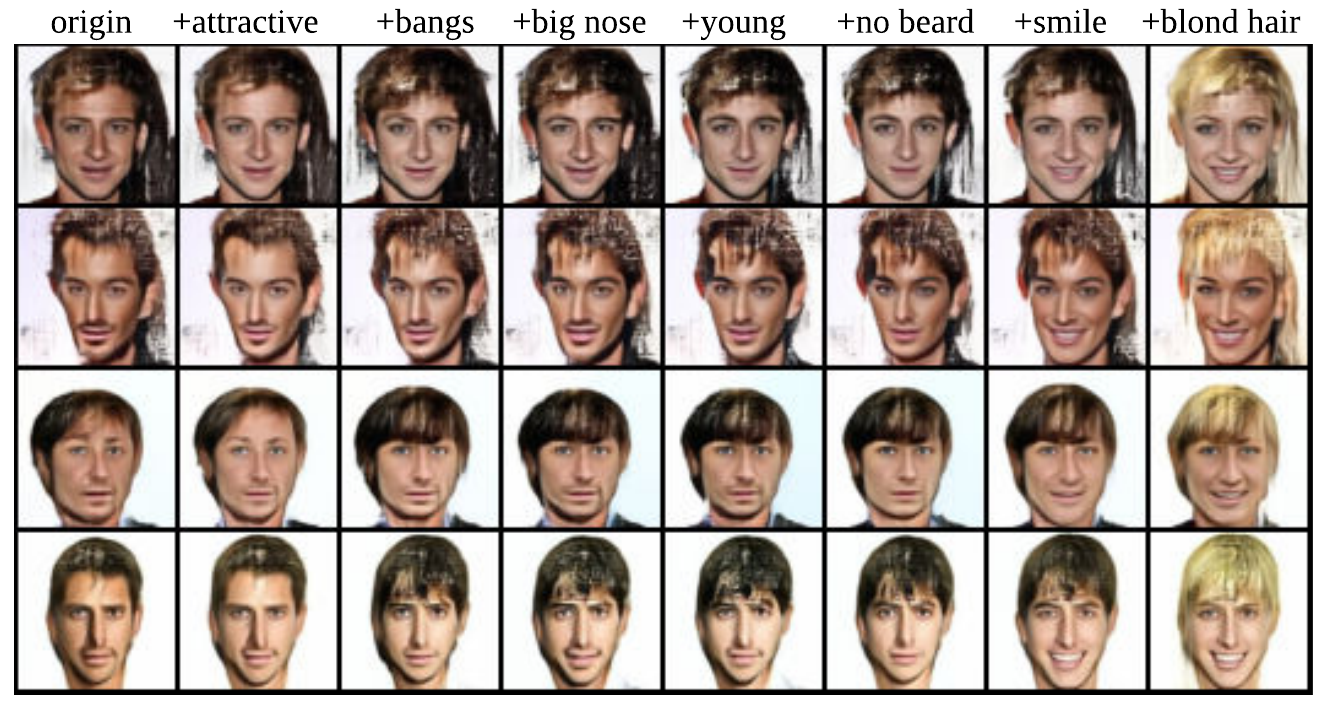}}
  \caption{Image synthesis under cumulative conditions demonstration. From left to right: adding different attributes step by step (specific attribute is annotated above the first row). (a) Images generated by regular Glow with pre-storing features. Identities and other attributes are interfered heavily; (b) Images generated by CAGlow with better controllability.}
  \label{fig:multi_attr}
\end{figure*}
\noindent \textbf{Image synthesis under cumulative conditions}. To further validate the controllability of our approach, we demonstrate the generation results of changing multiple attributes step by step. Because it is difficult for CGlow to change multiple attributes with identities persistent, we compare our approach with the regular Glow with pre-storing features. To maintain identities while changing attributes, the regular Glow must first parse all the latent vectors of original images and store a mean feature for each specific attribute. Then it infers the latent vector of an arbitrary image and changes this vector by adding pre-stored attribute feature, and thus it could generate targeted images with identity unchanged. This strategy works well when manipulating just one attribute. But the results are not ideal enough when manipulating multiple attributes. As shown in Figure~\ref{fig:multi_attr}, we add one more attribute to the original face images step by step. In regular Glow, adding the attributes `young'  and `no beard' causes the appearance of the attributes `makeup' and `long hair' and adding the attribute `blond hair' even results in the change of identity. In contrast, our model performs well by controlling the change of attributes independently under cumulative conditions.

Besides, our approach has two extra advantages over the regular Glow: 1) By a condition-latent-condition encoding-decoding strategy, we disentangle the feature of identities and attributes well, so it could produce non-interfered images; 2) We only feed some one-hot vectors of conditions into the encoder followed by the inverse flow to generate images, which does not need pre-storing attribute features and inference process for obtaining a specific latent vector, so CAGlow has a obvious improvement on time and space consumption. 

\begin{figure}
  \centering
  \includegraphics[width=0.95\linewidth]{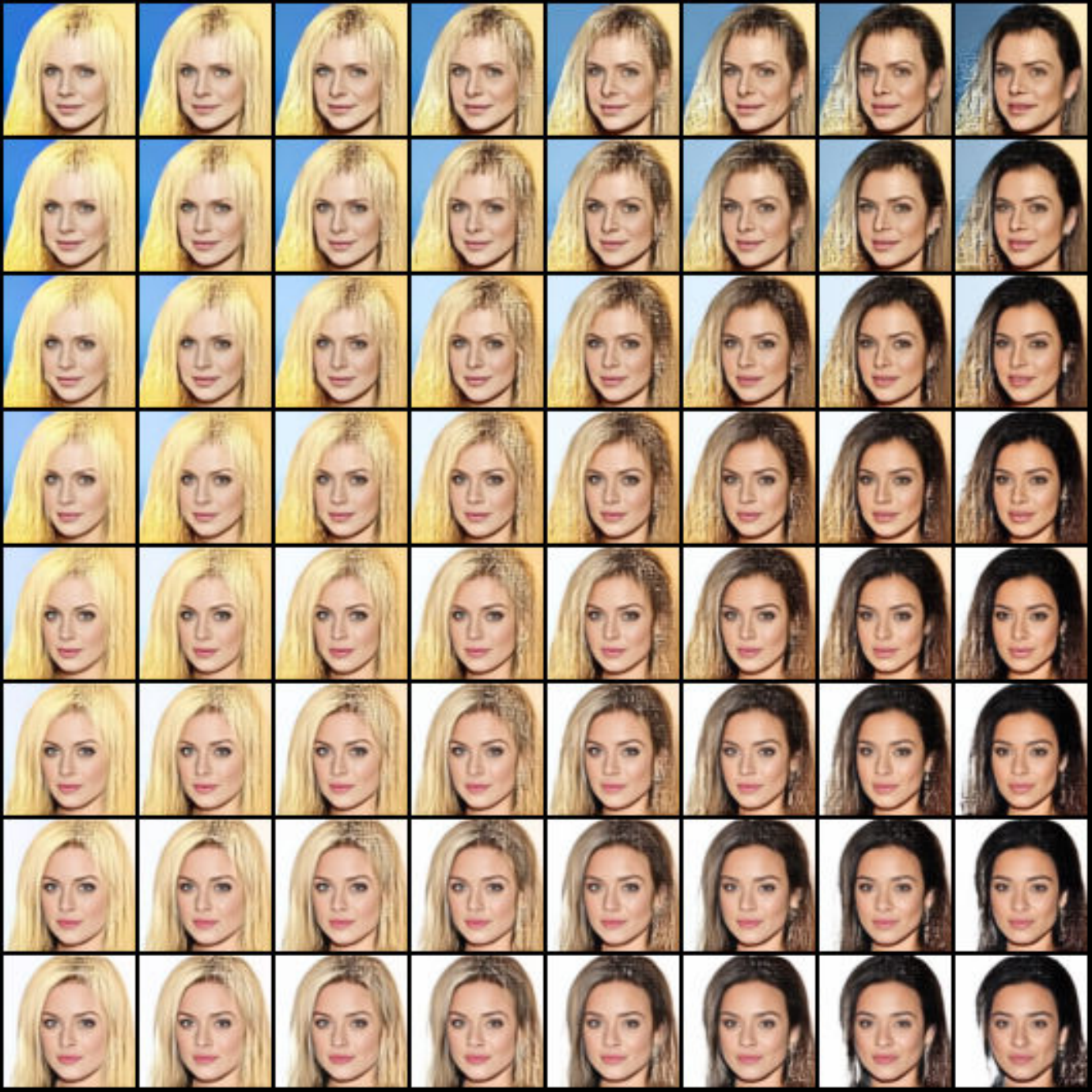}
  \caption{Interpolation both on ID and attribute. From top to bottom: interpolation on two different people. From left to right: interpolation on two different attributes (blond hair to black hair).}
  \label{fig:id_attr_interp} 
\end{figure}
\noindent \textbf{Smooth Interpolation}. We also demonstrate an interpolation generation results on two different identities and attributes simultaneously in Figure~\ref{fig:id_attr_interp}. The operation is completed by simply changing the input one-hot vector of two specific targets from $[0,1]$ to $[1,0]$. As one can see from the figure, the interpolation of one specific condition demonstrates continuous changes of generated images and has no negative impact on another condition. 

\subsection{Quantitative Comparisons}
In this part we perform some experiments to verify the superiority of our approach using some quantitative results. 

\begin{table}
\begin{center}
\begin{tabular}{|l|c|c|c|}
\hline
  & Glow & CGlow & CAGlow   \\
\hline
Acc (MNIST) & - & $98.89\%$ & $99.55\%$ \\
Acc (CelebA) & - & $87.43\%$ & $95.16\%$ \\
FID (MNIST) & $25.78$ & $29.64$ & $26.34$ \\
FID (CelebA) & $103.67$ & $126.52$ & $104.91$ \\
\hline
\end{tabular}
\end{center}
\caption{Accuracy and FID results on MNIST and CelebA. }
\label{table:acc_fid}
\end{table}
\noindent \textbf{Category preserving test}. We would take Fr\'echet Inception Distance (FID)~\cite{FID} and top-$1$ accuracy as our metric, to evaluate the realism, diversity and discriminability respectively. We pretrain the GoogLeNet~\cite{googlenet} on MNIST and CelebA dataset and then calculate the top-$1$ accuracy of the generated samples by different approaches. Following the method in~\cite{NIPS2017_7240}, we calculate the FID score of the generated samples on a pretrained GoogLeNet. As shown in Table~\ref{table:acc_fid}, our approach achieves better performance on both MNIST and CelebA dataset. The FID score of CAGlow is pretty close to the original Glow, which means that our approach could learn a good conditional distribution of latent vectors without losing diversity. 

\begin{table}
\begin{center}
\begin{tabular}{|l|c|c|c|c|}
\hline
Model & CGlow & CAGlow  \\
\hline
Accuracy & $87.36\%$ & $93.75\%$ \\
\hline
Variance & $0.0245$ & $0.0016$ \\
\hline
\end{tabular}
\end{center}
\caption{Accuracy of Attributes and Variance of AMP with different identities on CelebA.}
\label{table:acc_attr}
\end{table}
\noindent \textbf{Attribute preserving test}. Here we propose a novel evaluation metric Attribute Mean Probability (AMP) for testing the stability of the attributes. We first train $L$ different classifiers for $L$ different attributes based on CelebA dataset to obtain above $99\%$ precision. For any image $x_{i}$ with identity $i$, these classifiers could output the probabilities $p_{l}(x_{i})$ for different attributes $l\in\{1,...,L\}$. The value of AMP is calculated by $AMP_{i}=\frac{1}{L}\sum_{l=1}^{L}p_{l}(x_{i})$. Based on these classifiers, we could calculate the mean value of predicted accuracy for all generated samples and the variance of AMP along the identities. Our approach has better accuracy and lower variance, as reported in Table~\ref{table:acc_attr}.

\begin{table}
\begin{center}
\begin{tabular}{|l|c|c|c|}
\hline
\# Mpl & CGlow & Glow + pre-store & CAGlow  \\
\hline
1 & 0.004350 & 0.002245 & 0.000774 \\
2 & 0.023215 & 0.007213 & 0.002608 \\
3 & 0.047055 & 0.014352 & 0.005825 \\
4 & 0.077767 & 0.023767 & 0.009939 \\
\hline
\end{tabular}
\end{center}
\caption{The absolute value of the difference of AMP \textit{w.r.t.} the times of manipulation. Lower value means more stability.}
\label{table:camp}
\end{table}

\begin{figure*} 
  \centering
  \subfloat[Varying latent code for rotation.\label{disen_a}]{%
       \includegraphics[width=0.49\linewidth]{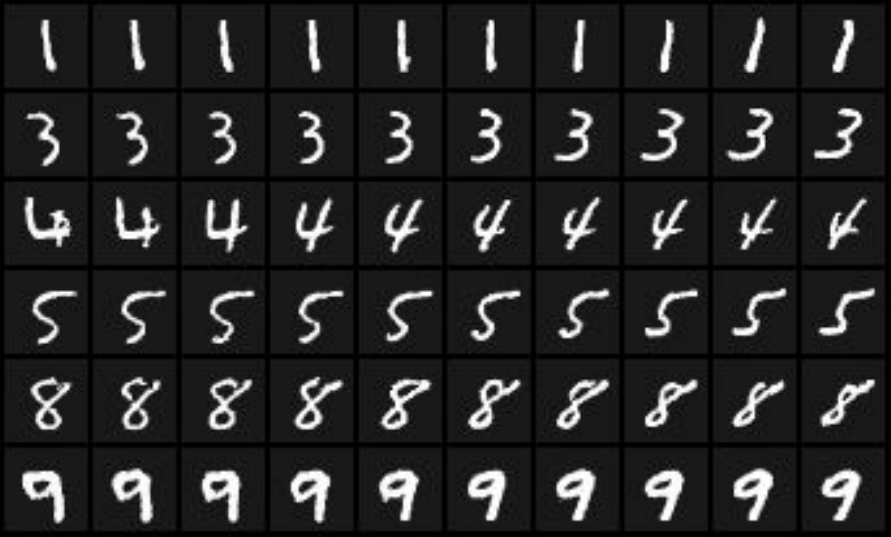}}
    \hfill
  \subfloat[Varying latent code for width.\label{disen_b}]{%
        \includegraphics[width=0.49\linewidth]{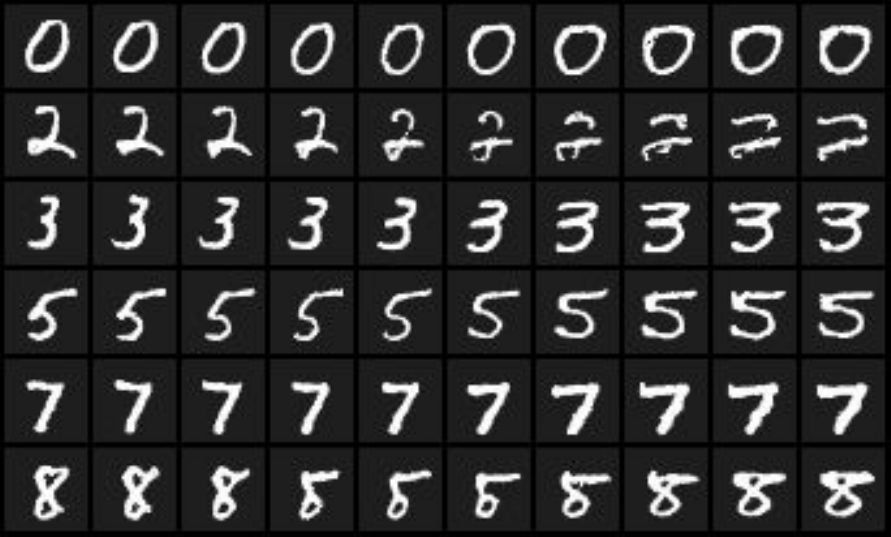}}
  \caption{Unknown properties exploration on MNIST~\cite{mnist}.}
  \label{fig:disentangle} 
\end{figure*}

\noindent \textbf{Cumulative conditions interfering test}. Same as the operation in~\ref{section42}, we change multiple attributes step by step. Based on $L$ pretrained classifiers mentioned above, we calculate the last-step and this-step mean probability of $L-1$ attributes except for the changed one for the generated images. Then we take the absolute value of the difference between the AMP of last step and this step as the metric. This evaluating metric describes the disturbing extent to the result. Smaller value means a more stable generating system and illustrates a better disentanglement between different attributes. We show that our results achieve the best performance, as summarized in Table~\ref{table:camp}.

\subsection{Interpretable Properties Exploration with Unsupervised Learning}

In the part we will explore some underlying properties in MNIST and CelebA dataset. Besides the conditional information these datasets provide, there are some unknown conditional information hidden. This experiment aims to demonstrate that our model could generate images conditioned on some unknown but interpretable properties. Note that these properties are found in an unsupervised manner. We do not add any supervision signals on the loss and only use an auto-encoding reconstruction loss for the input codes sampled from a prior distribution. We assume uniform distribution for unsupervised codes and take a mean square error loss for reconstruction. 

The results on MNIST are shown in Figure~\ref{fig:disentangle}. From this figure, we can see that the rotation direction and width of the generated digits changed continuously with the varying of the latent conditional codes respectively. 

\begin{figure} 
  \centering
  \includegraphics[width=0.99\linewidth]{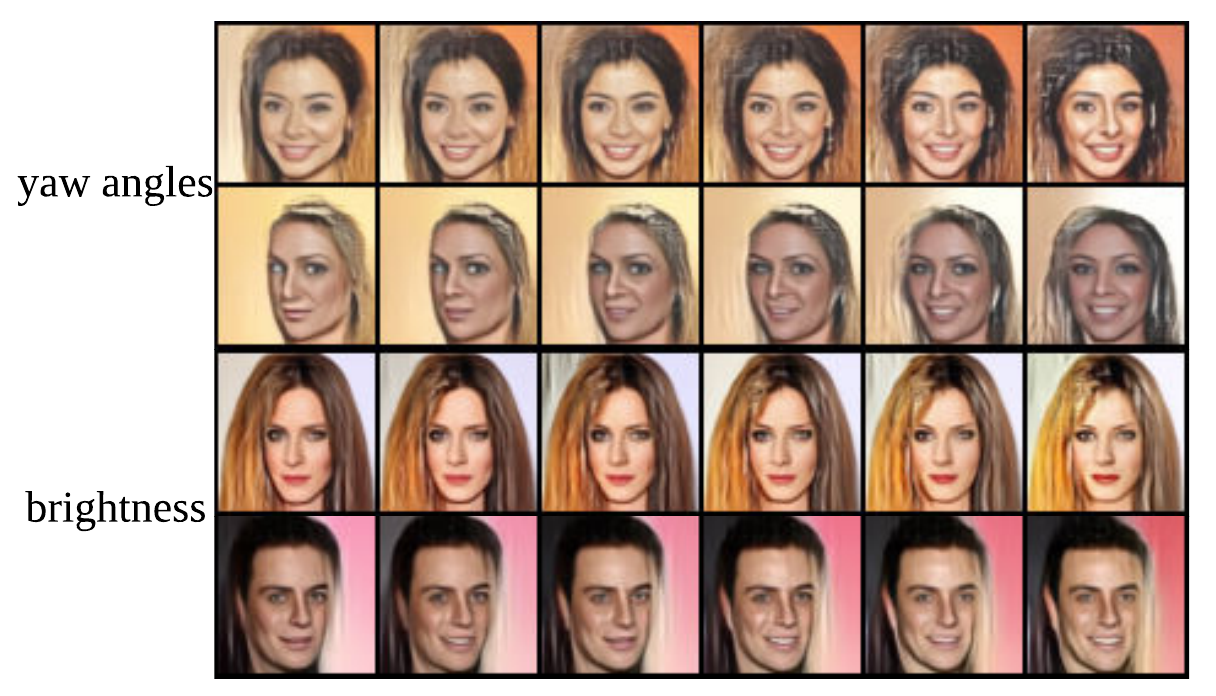}
  \caption{Unknown properties exploration on CelebA.}
  \label{fig:celeba_unsup} 
\end{figure}
We also show the exploration results on CelebA in Figure~\ref{fig:celeba_unsup}. As we can see, our approach could capture the underlying distribution for different yaw angles and brightness, which are not annotated in CelebA dataset. 


\section{Conclusion}
In this paper we proposed a novel generative model CAGlow that seamlessly unifies three sub-blocks: a reversible flow, an encoder and a supervision block and takes advantage of an adversarial training strategy. This framework provides great controllability and flexibility for synthesizing images conditioned on multiple annotations. Both qualitative and quantitative experimental results testified the superiority of the proposed approach to the vanilla version of Glow. In the future we plan to further investigate the impact of a more complex prior distribution instead of a simple Gaussian distribution on flow-based generative models. 

\noindent \textbf{Acknowledgement}. This work is supported in part by SenseTime Group Limited, in part by the General Research Fund through the Research Grants Council of Hong Kong under Grants CUHK14202217, CUHK14203118, CUHK14205615, CUHK14207814, CUHK14213616, CUHK14208417, CUHK14239816, and in part by CUHK Direct Grant. 

{\small
\bibliographystyle{ieee_fullname}

}

\end{document}